\PassOptionsToPackage{table,xcdraw}{xcolor}
\documentclass[sigconf,nonacm]{acmart}
\AtBeginDocument{%
  \providecommand\BibTeX{{%
    \normalfont B\kern-0.5em{\scshape i\kern-0.25em b}\kern-0.8em\TeX}}}

\setcopyright{acmlicensed}
\copyrightyear{2018}
\acmYear{2018}
\acmDOI{XXXXXXX.XXXXXXX}

\acmConference[CIKM '24]{Make sure to enter the correct
  conference title from your rights confirmation emai}{October 21--25,
  2024}{Woodstock, NY}
\acmISBN{978-1-4503-XXXX-X/18/06}

\setcopyright{none}
\renewcommand\footnotetextcopyrightpermission[1]{} 
\usepackage{listings}
\usepackage{tcolorbox}
\usepackage{caption}
\usepackage{subcaption}
\usepackage{tabularx}
\usepackage{multirow}
\usepackage{bm}

\usepackage{xcolor}
\usepackage{enumitem}

\begin{document}

\title{Text-centric Alignment for Multi-Modality Supervised Learning}


\author{Yun-Da Tsai}
\authornote{Both authors contributed equally to this research.}
\affiliation{%
  \institution{National Taiwan University}
  \country{}
}
\email{f08946007@csie.ntu.edu.tw}

\author{Ting Yu Yen}
\authornotemark[1]
\affiliation{%
  \institution{National Taiwan University}
  \country{}
}
\email{r11922042@ntu.edu.tw}

\author{Pei-Fu Guo}
\affiliation{%
  \institution{National Taiwan University}
  \country{}
}
\email{r12922217@ntu.edu.tw}

\author{Zhe-Yan Li}
\affiliation{%
  \institution{National Taiwan University}
  \country{}
}
\email{b09902111@ntu.edu.tw}

\author{Shou-De Lin}
\affiliation{%
 \institution{National Taiwan University}
 \country{}
}
\email{sdlin@csie.ntu.edu.tw}






\newcommand{\method}{TAMML }

\begin{abstract}
    This paper addresses the challenge of modality mismatch in supervised learning, where the modalities available during inference differ from those available during training. We propose an innovative method that utilizes Large Language Models with in-context learning and foundation models to enhance the generalizability of multimodal systems under these conditions. By leveraging the unique properties of text as a unified semantic space, this paper demonstrates significant improvements in handling unseen, diverse, and unpredictable modality combinations. The proposed solution not only adapts to varying modalities but also maintains robust performance, showcasing the potential of foundation models in overcoming the limitations of traditional fixed-modality frameworks in embedding representations. This study contributes to the field by offering a flexible, effective solution for real-world applications where modality availability is dynamic and uncertain.
\end{abstract}

\maketitle

\section{Introduction}
\label{sec:intro}



\begin{figure}[t]
    \centering
    \includegraphics[width=1.0\linewidth]{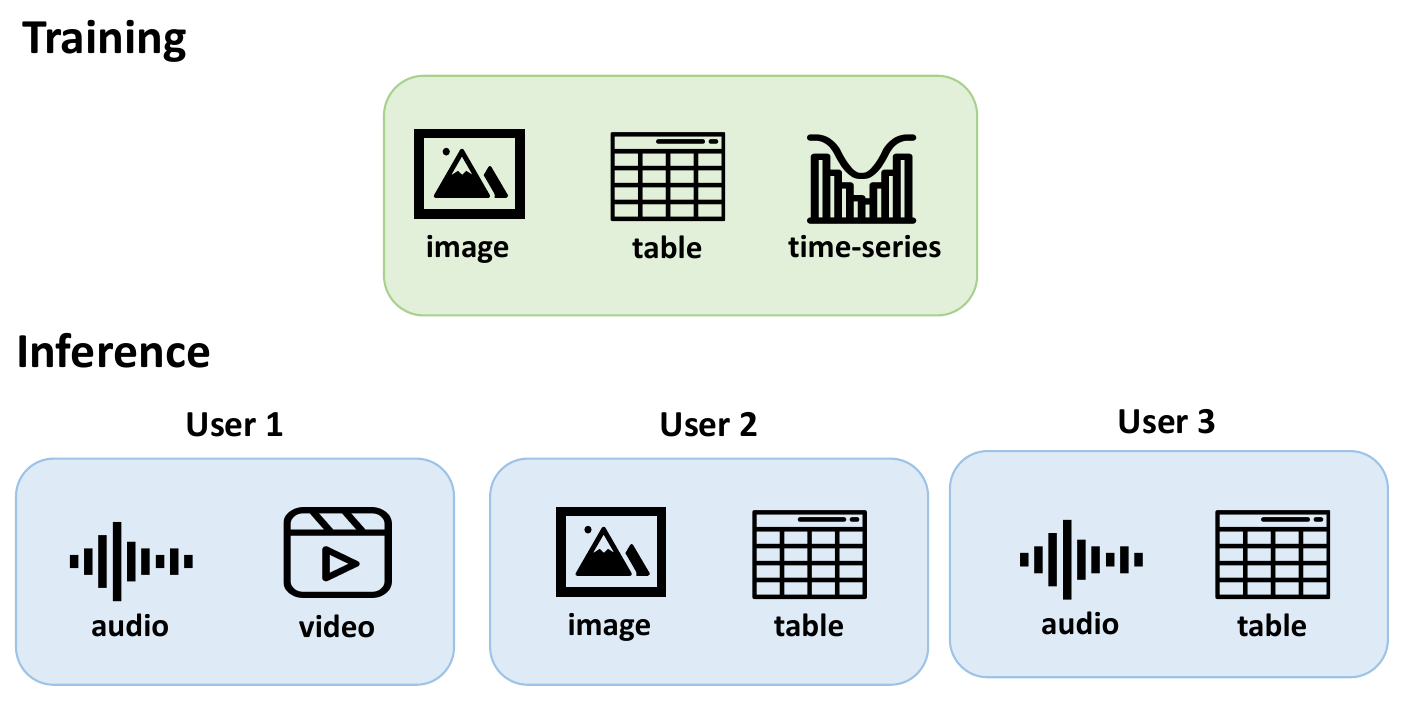}
    \caption{This paper establishes a general method for all mismatch types and combinations. As the figure shows, the model utilizes three modalities during training. Our unified model handles inference for any combination of modalities, such as User1’s unseen audio-video combination or the diverse combinations presented by User2 and User3.}
    \label{fig:intro}
\end{figure}


This work targets the challenge of developing a supervised learning model where the input modality in the testing (or prediction) phase differs from that in training.
The motivation for this research arises from the dynamic nature of real-world data, where modalities can unpredictably vary or even be absent at inference time. Consider the following scenarios:
1. A hospital has extensive image and text data about its patients, such as X-ray images and doctors' written diagnoses. This data can be used to train an AI model that diagnoses patients based on both image and text inputs. However, to enhance patient satisfaction, the hospital wants to develop a dialog system that can diagnose patients based on their audio descriptions of their symptoms. Typically, achieving this would require collecting audio data and employing transfer learning or domain adaptation techniques to align the information across modalities, a process that demands significant additional effort, cost, and time.
2. In the financial industry, a bank aims to train an AI assistant to recommend suitable financial products to customers. The existing data comprises millions of customers' past credit card transactions and purchase histories in the form of time series and tables. However, during the inference phase, the AI assistant must interact with customers using only text messages. Collecting sufficient text data to build a model to align text with time-series and transaction records can be expensive and resource-intensive.

Traditional multimodal learning methods, typically fixated on static modality combinations during both training and inference, fall short in such fluid environments. Therefore, this paper explores the possibility of creating a supervised model that utilizes only existing modalities (e.g., images and text in the hospital scenario, and time series and tables in the bank scenario) during training, yet allows for the incorporation of an unseen modality (e.g., audio signals or text inputs) during inference. If successful, this approach would eliminate the need for additional data collection and modality alignment, thereby reducing the associated costs and efforts.


There are two plausible directions to tackle this challenge.

First, we can rely on universal models pre-trained on vast datasets across numerous modalities to encode these modalities into embeddings. However, changes in the input modality necessitate retraining the discriminative downstream model for accurate predictions. From our experiments, we infer a classifier trained on embeddings from images and tabular data is not ideal for making inferences about audio inputs due to disparities in embedding dimensions and semantic content.

The second direction, which this paper adopts, is to convert every modality into a single modality and build the model based on that unified modality. We argue that converting all modalities into text could be a favorable choice. Text can serve as a unified semantic space, leveraging the extensive zero-shot prediction capabilities of Large Language Models (LLMs). The modality-invariant nature of text provides a versatile bridge across different data types, potentially circumventing issues like modality collapse and extending generalizability to unseen modalities. Furthermore, advanced text analysis tasks such as translation, summarization, and explanation have been extensively researched and integrated into LLMs, offering powerful capabilities to align various modalities effectively. 


Our objective is to develop a downstream model that is invariant across modalities. This model should be capable of being trained on data of certain modalities and performing zero-shot predictions on various modality combinations during testing, regardless of whether they were seen during training, as shown in Figure~\ref{fig:intro} and Figure~\ref{fig:overview}. Note that, as we will demonstrate through experiments, simply converting all modalities into text for training and inference is insufficient, as the mismatch between different modalities does not translate and align as seamlessly in text, especially in complex scenarios involving multiple modality combinations. To address these issues, our work has explored challenges such as modality alignment, translation, and augmentation. Our goal is to pioneer methods in multimodal learning that leverage text representations and in-context learning, establishing seamless integration across various modalities. Furthermore, we aim to explore the balance between flexibility and performance when employing text representations for multimodal learning, in contrast to traditional embedding approaches.

\method employs LLMs for data transformation across various modalities, with the aim of creating a unified semantic space. This process is conducted exclusively through in-context learning. Initially, we transform different modalities into text. Recently, various solutions have been developed, such as GPT-4~\cite{openai}, Blip2~\cite{li2023blip} for vision, and TabLLM~\cite{hegselmann2023tabllm} for tabular data. Following this, we engage LLMs in text-style translation across modalities, ensuring that all modalities in their textual representation adopt a consistent linguistic structure, thereby reducing the gap between different modalities. To further align these modalities within a closer semantic space, remove redundant information, and mitigate the heterogeneity inherent in text data from diverse sources, we further conduct modality summarization. This step involves a concise summarization of the translated data. Additionally, \method includes a reasoning augmentation step akin to the Chain-of-Thought~\cite{wei2022chain} method, where we enhance the data with LLMs to boost prediction and judgment capabilities. Moreover, we leverage LLMs as a source of large-scale external knowledge, enriching the data understanding and interpretative depth~\cite{KDA_2023}.
We aim to answer several hypotheses through extensive experiments. First, whether \method is more effective compared to existing solutions in predicting data of unseen modalities. Second, although this work focuses on predicting unseen modalities, we want to understand whether the proposed solution is effective when the modality in testing is already seen during training. Third, whether the text-as-the-medium strategy is more robust compared to embedding-based cross-modality transfer solutions.
We benchmarked \method against existing methodologies in closely related tasks, particularly focusing on zero-shot learning cross-modality translation, which involves translating unseen source data to a different target domain. Techniques like MIDiffusion~\cite{wang2023zero} and SDEdit~\cite{Meng2022SDEdit} demonstrate commendable performance in tasks such as domain translation within images. However, these methods encounter challenges when the source and target domains represent completely different modalities. 


Our contributions can be summarized as follows:
\begin{itemize}
    \item We investigate the potential advantage of using LLMs and text representation for multimodal learning. We propose \method, an in-context cross-modality translation method that utilizes foundation models to tackle training/testing modality mismatch and generalize to any unseen modality at test time.
    \item We demonstrate that \method can significantly outperform SOTA approaches by conducting multiple experiments on real-world datasets. We also have an ablation study to analyze the effectiveness of each component in \method.
    \item Additional experiments further verify that even when the testing modality is already seen during training, \method can still outperform the competitors by a large margin. 
\end{itemize}
The remainder of this paper is organized as follows: Section~\ref{sec:related} reviews preliminary works on Multimodal Foundation Models and Zero-shot Cross Modality Translation. Section~\ref{sec:method} introduces \method, detailing how we utilize LLMs to transform between different modalities and adapt to unseen inference modalities. Section~\ref{sec:experiments} presents our experimental results and demonstrates the effectiveness of \method, including ablation studies on the individual components of \method. Section~\ref{sec:analysis} provides further analysis, comparing text and embedding representations through visualization. The paper concludes with Section~\ref{sec:conclusion}, summarizing our findings and discussing limitations and future directions.

\begin{figure}[t]
    \centering
    \includegraphics[width=1.0\linewidth]{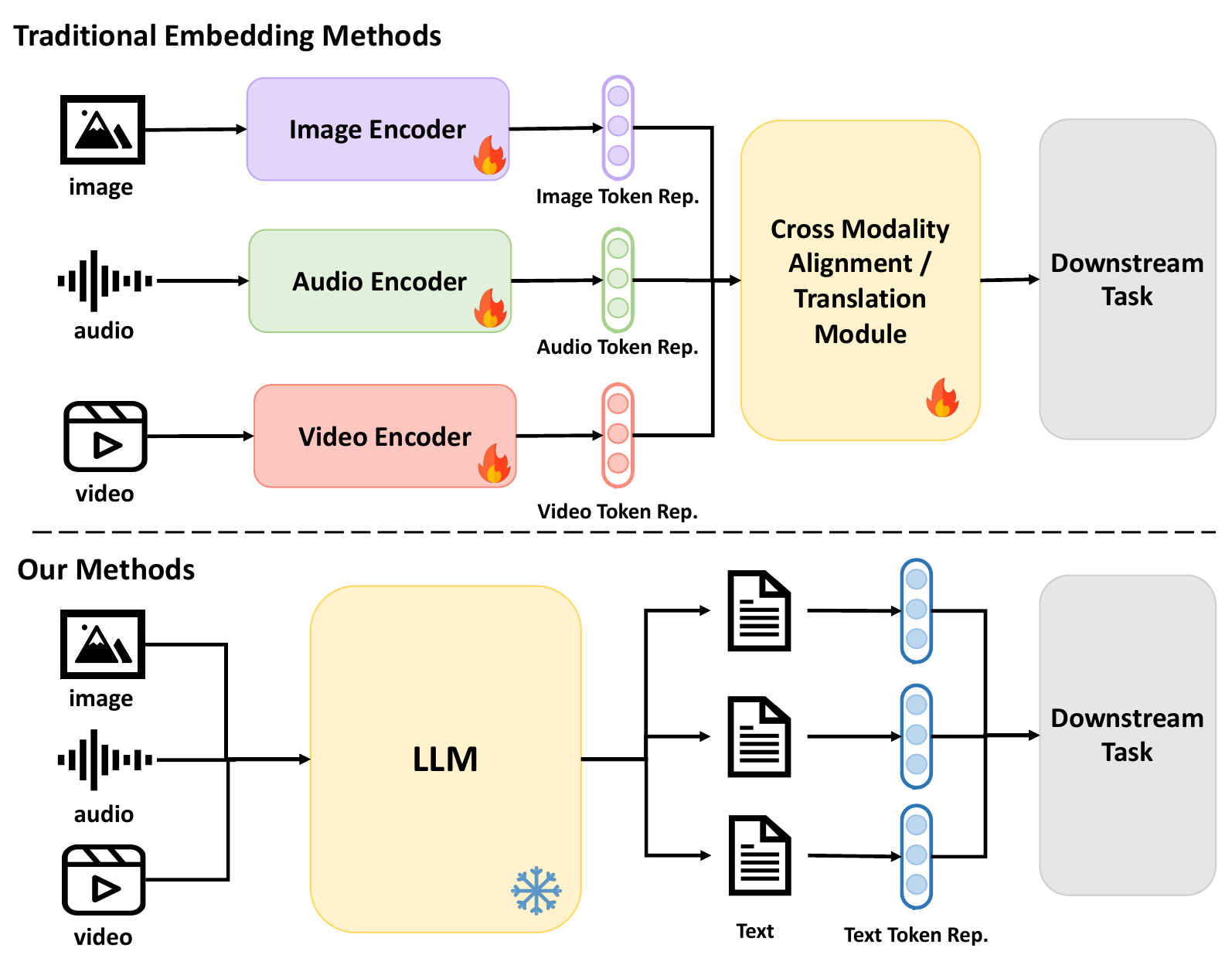}
    \caption{
    Traditional downstream training relies on embeddings extracted from upstream foundation models, with one foundation model designated for each modality. This approach limits the downstream model's ability to adapt to unseen modalities at test time without undergoing complete retraining. Previous research has addressed this issue by implementing zero-shot cross-modality translations during the inference phase.
    }
    \label{fig:embedding_based}
\end{figure}

\section{Related Works}
\label{sec:related}


\subsection{Multimodal Foundation Models}
\label{sec:related-foundation}
Recent advancements in large-scale foundation models have significantly enhanced content generation capabilities across various modalities. These developments span a wide range of applications, including text-to-image~\cite{ramesh2021zero,rombach2022high}, text-to-video~\cite{singer2022make}, audio-to-image~\cite{jamaludin2019you}, text-to-speech~\cite{ren2019fastspeech}, speech-to-gesture~\cite{ahuja2020style}, speaker-to-listener~\cite{ng2022learning}, language-to-pose~\cite{ahuja2019language2pose}, and even in the realms of speech and music generation~\cite{oord2018parallel}.
Nevertheless, aligning the semantic spaces of independently trained foundation models poses a significant challenge, hindering the ability of downstream models to seamlessly switch between upstream modalities at test time.

Extending beyond single-modality applications, large multimodal language models (MLLMs) have shown remarkable proficiency in both reasoning and generation tasks~\cite{yin2023survey}. For instance, Flamingo~\cite{alayrac2022flamingo} employs a vision encoder for feature extraction from images on top of the transformer backbone. On the other hand, Kosmos-2~\cite{peng2023kosmos} is comprised of a vision encoder, a linear projector, and a base transformer. These models directly concatenate image and text features, using unique tokens for differentiation.
However, despite the impressive capabilities of MLLMs in handling multimodal data, acquiring large-scale datasets encompassing multiple modalities remains a significant challenge.

Several works involve processing images with foundation models and combining the results into text for LLM training, linking visual information with text. For example, LLaVA~\cite{liu2023llava} utilizes GPT4 to transfer images into captions and object detection results into text descriptions as their multimodal training data. VideoChat-Text~\cite{li2023videochat} encodes video into textual descriptions. In medicine, OphGLM~\cite{gao2023ophglm} uses classification and segmentation models to extract information from fundus images and create diagnostic reports for LLMs. Similarly, ChatCAD~\cite{wang2023chatcad} transforms X-ray outputs from CAD models into natural language for LLM input.

\begin{figure*}[t]
    \centering
    \includegraphics[width=0.95\linewidth]{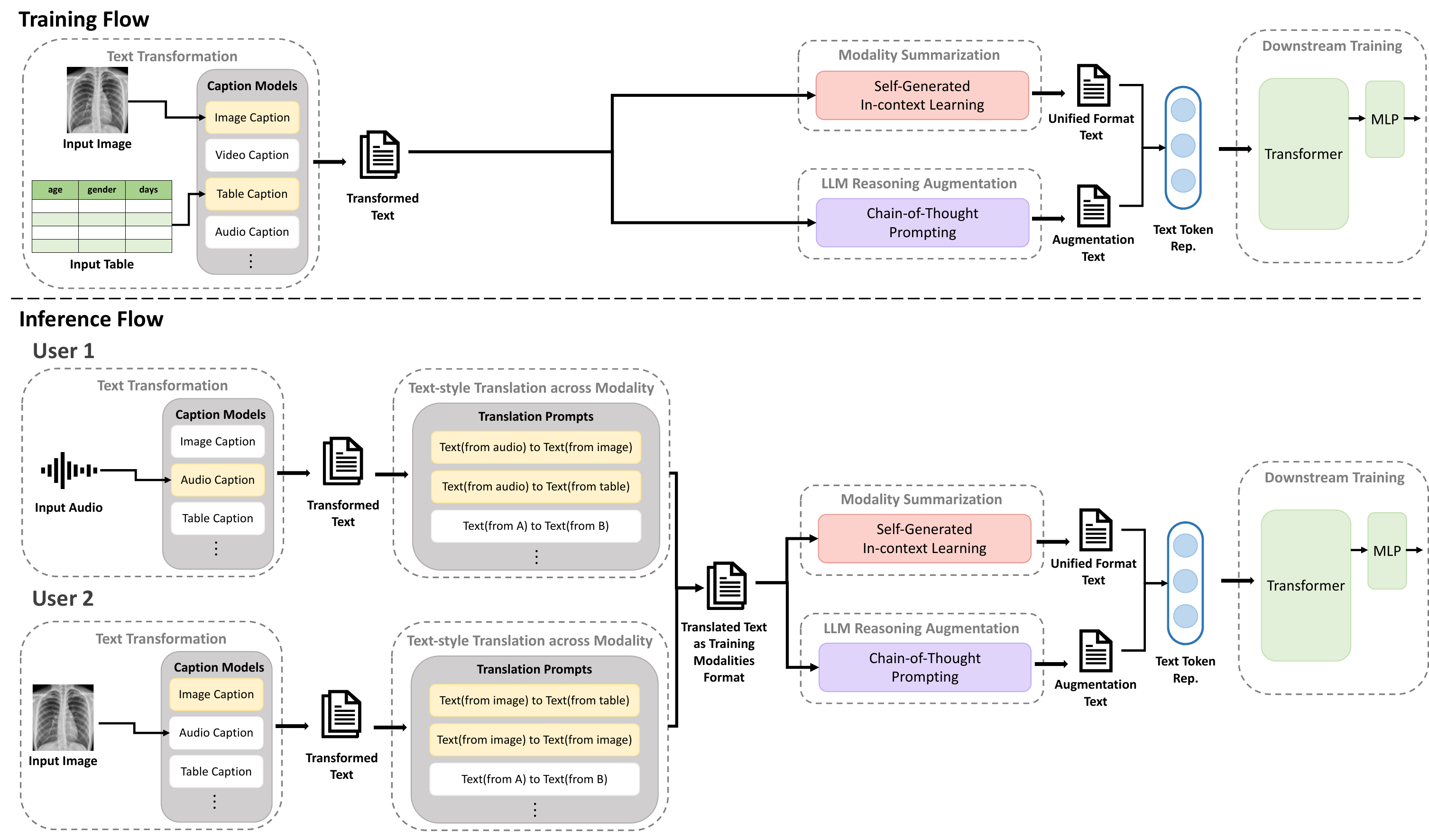}
    \caption{In the training phase, each raw input modality is transformed into text representations using a corresponding foundation model. Following the modality transformation, summarization, and augmentation are applied in parallel. Finally, the output texts are concatenated as the training inputs to a transformer model for downstream prediction. The inference phase follows a similar pattern, with the exception of utilizing an LLM for the text-style translation after the text transformation module. We apply a one-shot in-context learning approach to adapt the linguistic style as anticipated during training.}
    \label{fig:overview}
\end{figure*}

\subsection{Zero-shot Learning Cross Modality Translation}
\label{sec:related-modality-translation}

The challenge of cross-modality data translation without access to source modal data leads to a Zero-shot-Learning-based approach for this task. A fundamental difficulty with learning-based methods is their limited capability in handling unseen data classes~\cite{wang2021one,bucher2017generating,kuchibhotla2022unseen}. Zero-shot-Learning emerges as a robust strategy for scenarios where training and test classes do not overlap.
Traditional Zero-shot Learning methods often seek a direct projection from the image feature space to a semantic space. This is achieved through discriminative methods~\cite{palatucci2009zero,akata2015label} and generative models~\cite{long2017zero,wang2018zero}. In the realm of GAN-based zero-shot-learning for cross-modality data translation, models typically modify the latent representation of a pre-trained GAN, a process known as GAN inversion~\cite{zhu2020domain,shi2022semanticstylegan,brock2016neural,abdal2020image2stylegan++,da2022fast}. Perturbation-diffusion-based approaches to cross-modality data translation facilitate Zero-shot learning for these tasks~\cite{cheng2023adaptively,ho2020denoising,kawar2022denoising,song2020score}, showing impressive results when the numerical features of the source and target domains align~\cite{Meng2022SDEdit}. However, these methods can struggle when there is a significant disparity in the cross-domain appearance features.

\section{Methodologies}
\label{sec:method}

This section focuses on how \method enables training modalities to adapt to unseen testing modalities and unseen modality combinations. In Section~\ref{sec:method:problem}, we define the problem scenario and its notations, and Section~\ref{sec:method:transform} describes the individual text transformation methods for different modalities. Next, we introduce our main algorithm to connect information from multiple modalities. Section~\ref{sec:method:translation} discusses translating inference modalities to training modalities. Section~\ref{sec:method:summarization} summarizes different modalities into human-written descriptions, and Section~\ref{sec:method:reasoning} leverages external knowledge as data augmentation. Lastly, Section~\ref{sec:method:downstream} elucidates the flow of training and inference within the downstream task. The whole process is shown in  Figure.~\ref{fig:overview}.

\subsection{Problem Formalization}
\label{sec:method:problem}
Suppose we have a set \( M \) of \( p \) modalities, \( M = \{m_1, m_2, \ldots, m_p\} \). In the training phase, a subset of modalities \( M_T \subseteq M \) is used. In the inference phase, a different subset \( M_I \subseteq M \) is utilized. This subset meets the critical condition \( M_T \cap M_I = \emptyset \), ensuring no overlap in modalities between training and inference. 

Within this framework, we define two distinct datasets: one for the training phase and another for the inference phase. The training dataset \( D_T \) consists of \( n_T \) samples. Each sample \( x \) is restricted to \( M_T \), denoted as \( D_T = \{(x_{M_T}^i, y^i)\}_{i=1}^{n_T} \). Similarly, the inference dataset \( D_I \) consists of \( n_I \) samples, each restricted to \( M_I \), formalized as \( D_I = \{(x_{M_I}^i, y^i)\}_{i=1}^{n_I} \). Our algorithms are designed to build the model \( F \) on \( D_T \) and evaluate unseen data and modality combinations in \( D_I \). This evaluation measures the model’s ability to generalize knowledge in zero-shot multimodal learning.


\subsection{Text Transformation}
\label{sec:method:transform}
For the challenge of the heterogeneity between different modalities, we attempt to map the data of different modalities into a similar semantic space at the raw input stage, integrating modalities with minimal processing. Compared to traditional modality learning approaches that require using embedding features and need to train a new embedding layer for each task or new modality which challenges in zero-shot scenarios. Our approach simplifies this by using pre-trained MLLM or rule-based methods to convert raw inputs into text, enhancing efficiency in cross-modality applications. 

For image modality, we utilize the state-of-the-art image caption model, which allows us to take images and answer questions about them. In this stage, the image caption model is prompted to describe images and generate detailed text output, effectively translating visual content into descriptive text. For table modality, we implement the simple text serialization template, which is "The column name is values," proposed by TabLLM\cite{hegselmann2023tabllm}.
According to TabLLM's study, this template is proven to outperform zero-shot tabular learning in the LLM. For text modality, we keep human-written sentence forms without conversion to preserve the natural linguistic structure. After transformation, the sentence order of each sample \( x \)  is determined by the order of modalities in \( M \) to concatenate the text sentences and is used as the input for the next step. We have included and compared various SOTA image caption models in Table~\ref{table:average_model_capability}.

\noindent \textbf{A Real-world Example}: When predicting diseases, we often have access to patients' pathology table reports, medical imaging, and audio of patient narration. First, we will perform text transformation on these data. For the images, we use an image caption model that outputs descriptions such as "\textit{The patient has sigmoid colon cancer causing an obstruction, which has led to dilation in the descending colon.}" For the tables, we transform data with the template and output statements like "\textit{Histologic Type is Adenocarcinoma}" and "\textit{Histologic Grade is Moderately differentiated.}" For the audio files, we use the audio caption model that outputs descriptions such as "\textit{I've been a little bloated for two weeks, and I have had only three bowel movements.}"

\subsection{Text-style Translation across Modality}
\label{sec:method:translation}

After text transformation, we obtain textual representations for each modality. During the inference phase, we employ LLMs to perform text-style translation across different modalities, reducing the mismatch between training and inference data.
This is because, although all types of data are converted into textual representations, there are still syntactic and semantic gaps between the transformed text across different modalities, which are highly dependent on the foundation models used for transformation. For instance, the textual representations of tabular data can follow a static template, while the textual representation of the image is affected by the language style of the image caption model. This divergence in textual representations leads to gaps in semantic space distribution. Our solution is to adopt in-context learning in LLMs to translate the text style to the one in the target space. Text-style translation function \(T \) can be formulated as follows:
\[ T : M_I \rightarrow M_T \]

Specifically, we enhance the capability of the LLMs to effectively interpret the semantic content and convert the modality text from  \(M_I \) format into  \(M_T \) format. This transformation is achieved through few-shot in-context learning, where LLMs are exposed to three examples, which are samples from  \(D_T \), illustrating the desired translation for our modality combination used in training and inference. Moreover, to mitigate information loss during conversion, we specifically emphasize maintaining the integrity of the original information in our prompt. This emphasis ensures that the essential information content and cross-modal connections are preserved during the cross-modal translation process.   Figure.~\ref{fig:prompts} demonstrates the details of the Text-style translation prompt.

\noindent \textbf{A Real-world Example}: When the training combination for disease prediction includes table modality, and only video modality data is available at inference, we will perform text-style translation on the textual representation of audio data. Continuing the example from Section~\ref{sec:method:transform}, the textual representation of the audio, "\textit{I've been a little bloated for two weeks, and I have had only three bowel movements},", is translated as "\textit{Symptom is Bloating. The symptom is difficulty with bowel movements. Duration is Two weeks.}"


\subsection{Modality Summarization}
\label{sec:method:summarization}

Compared with text-style translation, which maps the inference distribution directly to the training distribution, modality summarization involves utilizing an LLM to summarize the translated data. This step aims to achieve two main objectives: (1) It facilitates the alignment of different modalities in a closer semantic space. As discussed in the previous section, varying text transformation methods cause gaps in semantic space. \method further generalizes the text representation across all modalities with similar linguistic styles and formats. (2) It identifies the connections and interactions among elements of different modalities. This process enhances information quality by promoting interaction, such as generating novel information, highlighting shared core information, and eliminating redundant information.

The modality summarization is also generated by LLMs. Our experiment is divided into two steps. First, we extract a sample in  \(D_T \) with the inference modality combination. We pre-define the linguistic styles in the prompt, instructing the LLM to integrate information across diverse modalities to produce a summarized output. Second, the output is merged back into our original prompt to form a demonstration, and then one-shot in-context learning is applied to each sample x in \(T(D_T)\). Each sample may follow the textual structure of the demonstration established in the first step, aiming to reduce heterogeneity.  Figure.~\ref{fig:prompts} demonstrates the details of the modality summarization prompt.

\noindent \textbf{A Real-world Example}: Building on the example from Section~\ref{sec:method:transform}, now the input includes two modalities: image and table. We summarize the textual representations from these modalities. Here is how the summarization looks: "\textit{The patient has moderately differentiated adenocarcinoma of the sigmoid colon, causing an obstruction and dilation of the descending colon.}"

\subsection{LLM Reasoning Augmentation}
\label{sec:method:reasoning}
We apply LLMs in terms of reasoning similar to Chain-of-Thought~\cite{wei2022chain} method and leverage LLMs as a source of large-scale external knowledge similar to ~\cite{KDA_2023} to achieve data augmentation.

\begin{figure*}[t]
    \centering
    \includegraphics[width=0.8\linewidth]{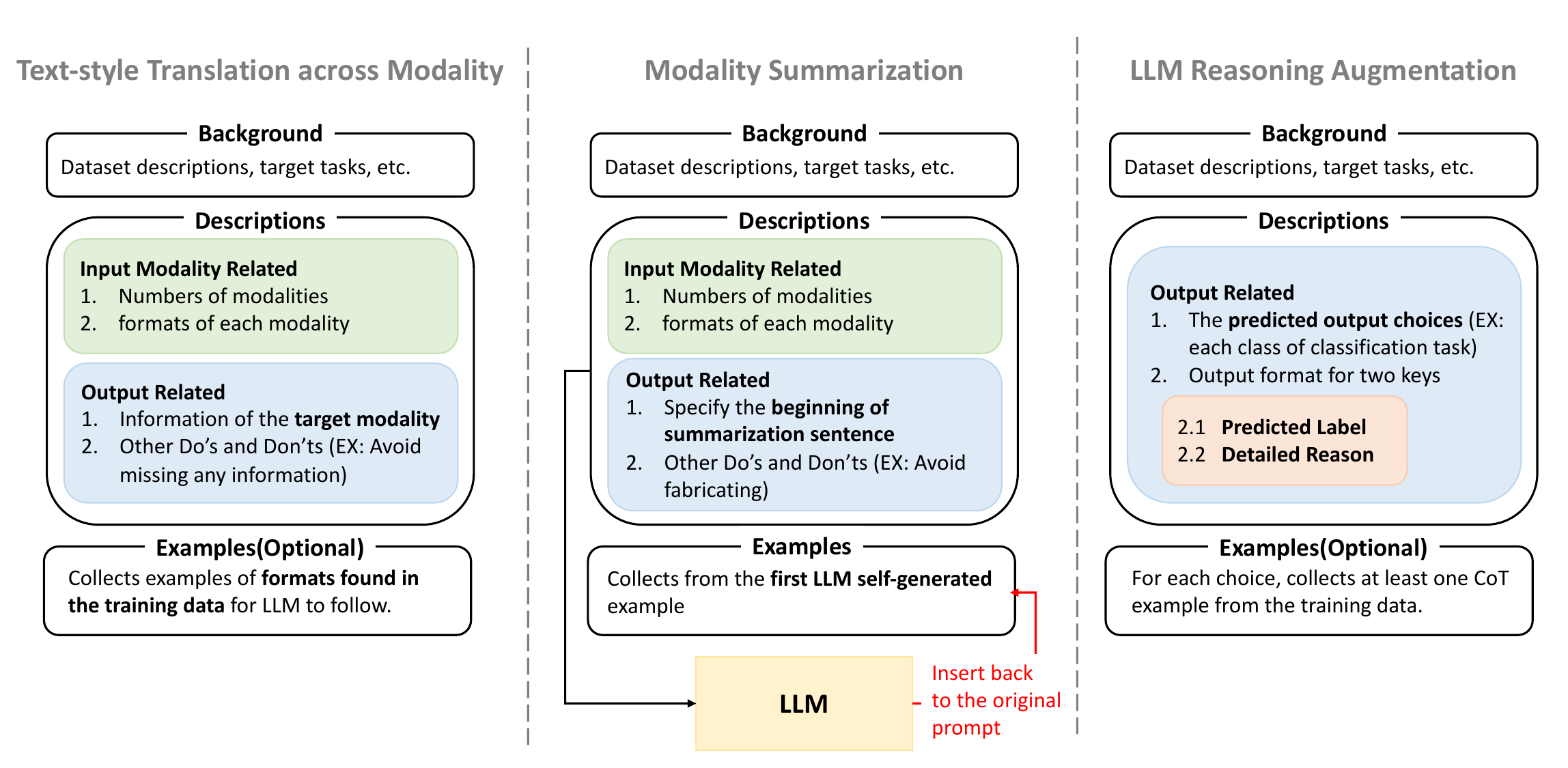}
    \caption{Examples of prompt templates for each modules}
    \label{fig:prompts}
\end{figure*}

First, the LLM is assigned a specific prediction task accompanied by straightforward instructions and examples to leverage its external knowledge while analyzing the concepts and information present in the original textual inputs. Throughout the task, the LLM will make predictions and provide comprehensive explanations for each input sample. By utilizing the outputs derived from this predictive reasoning process, we enhance the original input information and achieve data augmentation.
Next, for each reasoning process, the LLM is furnished with pertinent background information and examples related to its prediction tasks.
It is subsequently assigned an assistant role to provide informative reasoning and accurately predict either objective values in regression tasks or labels in classification tasks.  Figure.~\ref{fig:prompts} demonstrates the details of the reasoning augmentation prompt.

\noindent \textbf{A Real-world Example}: Building on the example from Section~\ref{sec:method:transform}, now the input includes two modalities: image and table. The current goal is to determine whether a patient requires hospital observation. The results after augmentation are as follows: "\textit{The obstruction in the sigmoid colon can lead to increased risks of bowel perforation, where the colon wall might rupture due to increased pressure. This complication is serious and requires immediate medical intervention.}"

\subsection{Downstream Training}
\label{sec:method:downstream}
This section elucidates the entire flow of training and inference within downstream tasks. The process is shown in  Figure.~\ref{fig:overview}. Initially, the raw input contains different modalities, each of which is very heterogeneous in expression. First, we transform each modality into text. In the inference phase, we conduct the additional text-style translation step. In parallel, the text representation is then subjected to two distinct yet complementary processes: modality summarization and reasoning augmentation. The summarization aligns textual representations in diverse modalities, while the reasoning augmentation step integrates these points with external knowledge, thereby enhancing the overall informational context. 

The transformed text representation is then fed into a transformer. More specifically, we adopt Longformer\cite{beltagy2020longformer} as the transformer architecture. After mean pooling is applied to aggregate the contextual representations, a Multilayer Perceptron (MLP) is employed to process the pooled features and output the predictions.
The loss function employed is contingent on the downstream task. In this paper, we use Cross Entropy Loss for classification tasks and Mean Squared Error for regression tasks.



\section{Experiments}
\label{sec:experiments}

Here, we articulate our hypotheses and address the research questions to evaluate the effectiveness of \method. Q1: Under modality mismatch scenarios, is \method better than the embedding-based SOTA solutions? Q2 (follow Q1): Is \method still effective for situations in which the testing modality has been involved during training? (i.e. training: all modalities, testing: some of the modalities) Q3: Is text representation generally more robust than embedding representation for cross-modality translation?

\noindent\textbf{Setup} In this section, we primarily present results from GPT-4-Vision for image captioning, unless specified otherwise. For additional results involving other image caption models, please refer to Table~\ref{table:average_model_capability}. Furthermore, our performance enhancement is all calculated using the formula: (measured performance - baseline performance) / baseline performance.

\subsection{Q1: Under Modality Mismatch Scenarios, How Does \method Compare To the SOTA?}
\label{sec:exp:q1}

In this section, we focus on modality mismatch scenarios where training and testing modalities are completely different. We mainly compare our results to several zero-shot cross-modality data translation methods.

\subsubsection{Dataset}\label{sec:exp:q1:dataset}
\textit{PetFinder.my Adoption Prediction}~\cite{petfinder-adoption-prediction} examines what factors predict how quickly a pet is adopted after being listed. The dataset is a composite of the following modalities:  
\begin{itemize}[leftmargin=*]
    \item Text: contains the description of the status of the pet
    \item Image: contains a profile photo of the pet
    \item Tabular: contains basic information, such as gender and breed. 
\end{itemize}

\noindent\textit{Airbnb Pricing Prediction}~\cite{insideairbnb} is composed of the following modalities used for making a regression prediction of housing prices:
\begin{itemize}[leftmargin=*]
    \item Text: contains the human-written description of the homestay, the neighborhood description, and the host's profile.
    \item Image: contains images of the homestay
    \item Tabular: delivers essential details such as location, rating score, and review counts.
\end{itemize}

\noindent\textit{Avito Demand Prediction}~\cite{avito-demand-prediction} predicts the likelihood of an ad selling something based on user item and context features:
\begin{itemize}[leftmargin=*]
    \item Text: contains the ad title and description.
    \item Image: contains a profile photo of the item.
    \item Tabular: contains basic information, such as region, city, item category, etc.
\end{itemize}

\begin{table*}[t]
    \centering
    \setlength\tabcolsep{10pt}
    \resizebox{\textwidth}{!}{%
    \begin{tabular}{ll|lll|lllll}
    \toprule
         \multirow{2}{*}{Training}              &       \multirow{2}{*}{Testing}       &         \multirow{2}{*}{SDEdit} &  \multirow{2}{*}{DDRM}  & \multirow{2}{*}{Idinvert} &  \multicolumn{5}{c}{\textbf{TAMML Methods}}                      \\
         ~              &       ~       &          ~                    &        ~              &     ~                   &    \textbf{LLaMa 3 8B} & \textbf{Mistral 7B} & \textbf{Mixtral 8x7B} & \textbf{Mixtral 8x22B} & \textbf{GPT-3.5} \\
    \midrule
        \rowcolor{gray!30}
        \multicolumn{10}{c}{\textbf{{PetFinder | Accuracy $\uparrow$}}} \\
    \hline
        text+image     & tabular       & 0.282  & 0.291 & 0.279    & 0.309      & 0.301      & 0.317        & 0.332         & \textbf{0.348}   \\
        text+tabular   & image         & 0.289  & 0.277 & 0.286    & 0.306      & 0.322      & 0.335        & 0.323         & \textbf{0.380}   \\
        image+tabular  & text          & 0.281  & 0.297 & 0.279    & 0.303      & 0.304      & 0.320        & 0.313         & \textbf{0.355}   \\
        text           & image+tabular & 0.291  & 0.283 & 0.289    & 0.315      & 0.304      & 0.330        & \textbf{0.368}         & 0.344   \\
        text           & image         & 0.289  & 0.276 & 0.287    & 0.355      & 0.341      & 0.307        & 0.360         & \textbf{0.374}   \\
        text           & tabular       & 0.293  & 0.259 & 0.277    & 0.295      & 0.286      & 0.325        & 0.341         & \textbf{0.357}   \\
        image          & text+tabular  & 0.290  & 0.297 & 0.284    & 0.310      & 0.322      & \textbf{0.346}        & 0.342         & 0.341   \\
        image          & text          & 0.288  & 0.282 & 0.280    & 0.323      & 0.325      & 0.329        & \textbf{0.330}         & 0.319   \\
        image          & tabular       & 0.291  & 0.287 & 0.284    & 0.322      & 0.302      & 0.341        & 0.319         & \textbf{0.348}   \\
        tabular        & text+image    & 0.290  & 0.271 & 0.285    & 0.314      & 0.309      & 0.327        & 0.333         & \textbf{0.360}   \\
        tabular        & text          & 0.289  & 0.265 & 0.280    & 0.295      & 0.294      & 0.302        & 0.317         & \textbf{0.364}   \\
        tabular        & image         & 0.289  & 0.263 & 0.277    & 0.294      & 0.305      & 0.338        & 0.311         & \textbf{0.364}   \\
    \midrule
        \multicolumn{2}{c|}{Average} & 0.289 & 0.279 & 0.282 & 0.312 & 0.310 & 0.326 & 0.332 & \textbf{0.355}\\
    \midrule[\heavyrulewidth]
        \rowcolor{gray!30}
        \multicolumn{10}{c}{\textbf{{Airbnb | MSE $\downarrow$}}} \\
    \hline
        text+image     & tabular       & 0.935  & 0.600 & 0.799    & 0.303      & 0.371      & 0.326        & \textbf{0.313}         & 0.367   \\
        text+tabular   & image         & 0.656  & 0.778 & 0.643    & 0.626      & 0.466      & 0.451        & \textbf{0.447}         & 0.508   \\
        image+tabular  & text          & 0.514  & 0.565 & 0.781    & 0.413      & 0.325      & \textbf{0.312}        & 0.359         & 0.332   \\
        text           & image+tabular & 1.548  & 0.914 & 0.915    & 0.315      & 0.368      & 0.323        & \textbf{0.284}         & 0.421   \\
        text           & image         & 1.513  & 0.895 & 1.010    & 0.537      & 0.521      & 0.439        & \textbf{0.404}         & 0.520   \\
        text           & tabular       & 1.061  & 0.824 & 0.931    & 0.308      & 0.348      & 0.345        & \textbf{0.297}         & 0.448   \\
        image          & text+tabular  & 0.556  & 0.530 & 0.602    & 0.431      & \textbf{0.368}      & 0.382        & 0.392         & 0.395   \\
        image          & text          & 0.678  & 0.589 & 0.759    & 0.439      & 0.389      & \textbf{0.375}        & 0.421         & 0.391   \\
        image          & tabular       & 0.592  & 0.538 & 0.516    & 0.459      & 0.452      & 0.487        & \textbf{0.405}         & 0.414   \\
        tabular        & text+image    & 0.637  & 0.675 & 0.662    & 0.467      & 0.347      & 0.310        & 0.379         & \textbf{0.280}   \\
        tabular        & text          & 0.569  & 0.693 & 0.707    & 0.481      & 0.341      & 0.313        & 0.339         & \textbf{0.301}   \\
        tabular        & image         & 0.609  & 0.715 & 0.615    & 0.627      & 0.461      & \textbf{0.431}        & 0.535         & 0.551   \\
    \midrule
        \multicolumn{2}{c|}{Average} & 0.822 & 0.693 & 0.745 & 0.451 & 0.396 & \textbf{0.375} & 0.381 & 0.411\\
    \midrule[\heavyrulewidth]
        \rowcolor{gray!30}
        \multicolumn{10}{c}{\textbf{{Avito | MSE $\downarrow$}}} \\
    \hline
        text+image     & tabular       & 0.103  & 0.113 & 0.126    & 0.045      & 0.045      & 0.043        & \textbf{0.041}         & 0.044   \\
        text+tabular   & image         & 0.130  & 0.133 & 0.142    & 0.048      & 0.048      & 0.047        & 0.048         & \textbf{0.046}   \\
        image+tabular  & text          & 0.113  & 0.125 & 0.137    & 0.045      & 0.045      & 0.045        & \textbf{0.043}         & 0.046   \\
        text           & image+tabular & 0.124  & 0.123 & 0.131    & 0.046      & 0.047      & 0.046        & 0.046         & \textbf{0.045}   \\
        text           & image         & 0.124  & 0.122 & 0.129    & 0.048      & 0.050      & \textbf{0.047}        & 0.048         & \textbf{0.047}   \\
        text           & tabular       & 0.127  & 0.124 & 0.134    & 0.045      & 0.046      & 0.046        & 0.046         & \textbf{0.044}   \\
        image          & text+tabular  & 0.123  & 0.126 & 0.134    & \textbf{0.044}      & \textbf{0.044}      & \textbf{0.044}        & \textbf{0.044}         & \textbf{0.044}   \\
        image          & text          & 0.118  & 0.124 & 0.129    & \textbf{0.045}      & 0.046      & 0.047        & 0.046         & \textbf{0.045}   \\
        image          & tabular       & 0.119  & 0.126 & 0.134    & 0.044      & 0.044      & 0.045        & \textbf{0.043}         & 0.044   \\
        tabular        & text+image    & 0.128  & 0.139 & 0.137    & 0.046      & 0.045      & 0.046        & \textbf{0.044}         & 0.046   \\
        tabular        & text          & 0.124  & 0.131 & 0.138    & 0.046      &\textbf{ 0.044}      & 0.047        & \textbf{0.044}         & 0.045   \\
        tabular        & image         & 0.126  & 0.137 & 0.140    & 0.050      & 0.047      & 0.048        & \textbf{0.046}         & 0.048   \\
    \midrule
        \multicolumn{2}{c|}{Average} & 0.122 & 0.127 & 0.135 & 0.046 & 0.046 & 0.046 & \textbf{0.045} & \textbf{0.045} \\
    \bottomrule
    \end{tabular}}
    \caption{This table presents a detailed comparison, highlighting \method's performance against all baseline models under modality mismatch scenarios. The PetFinder dataset uses accuracy as the key evaluation metric. The Airbnb dataset and the Avito dataset both use Mean Squared Error (MSE) as the key evaluation metric.}
    \label{tab:main}
\end{table*}

\subsubsection{Large Language Models}\label{sec:exp:q1:models}
To validate our framework and demonstrate its capability to enhance performance, we experiment with different LLM foundation models, including GPT3.5, and open-source models of different sizes in our architecture.

\subsubsection{Competitors}\label{sec:exp:q1:competitors}
Here, we adopt the embedding-based cross-modality translation solutions. A high-level illustration of conventional embedding-based downstream training is demonstrated in Figure~\ref{fig:embedding_based}.
In our experiments, we applied various foundation models, each designated for different modalities, as encoders that extract embedding representations. Afterward, alignment layers are used for learning and concatenating each representation.
Following these modality encoders, we concatenate the representative outputs from all modalities and feed them into a transformer encoder. After fine-tuning, the model generates cross-modality alignment outputs.

In the experiments, we compared \method to several zero-shot cross-modality data translation methods built on top of embedding-based downstream model training.
We include two perturbation-diffusion-based methods, SDEdit~\cite{Meng2022SDEdit}, DDRM ~\cite{kawar2022denoising}, and one GAN inversion method, In-domain inversion GAN (Idinvert)~\cite{zhu2020domain}.
All these methods include training a generative model (diffusion or a GAN) that can translate the input embedding distribution into a distribution similar to training modalities.
At test time, we use the trained generative models to generate synthesized modality embeddings (similar to the distribution of training modalities) for various unseen modalities.
For diffusion models, we train a score-based generative model~\cite{ho2020denoising} and a backbone model for each modality combination. In our implementation, we leverage DDIM~\cite{song2020denoising} as the pre-trained model for both SDEdit and DDRM.
For GANs, we leverage the StyleGAN~\cite{karras2019style} implementation as the backbone model.

\subsubsection{Results}
The key findings outlined in Table~\ref{tab:main} underscore the superior performance of \method, which achieves substantial gains over competing baselines across various modality combinations and different foundation models. These results suggest that the proposed strategy, which integrates LLMs' in-context learning with foundation models, holds a decisive edge over all existing methods. Specifically, with the best-performing GPT-3.5 on the PetFinder dataset, \method enhances accuracy by an average of approximately 21\%, significantly outperforming the best-performing baseline methods. Similarly, in the Airbnb dataset, \method achieves an average reduction in mean square error of around 54\%, dwarfing the maximum 16\% error reduction seen with alternative baselines. Further examination of the differences among various foundation models within the \method framework underscored the impact of model size on quality. For instance, Mixtral 8x22B improved accuracy by 7\% on the PetFinder dataset compared to Mistral 7B. For complex tasks such as summarization and translation, larger models performed better. However, even smaller models showed improvement compared to baselines in mismatch scenarios.

\subsection{Q2: Is the Proposed Solution Still Effective when there is no modality mismatching?}
\label{sec:exp:q2}

This section presents experiments designed to assess the performance of \method given all modalities are available during training. This setup evaluates whether \method can surpass embedding-based solutions even in scenarios without modality mismatches. In this section, our \method framework employs the Mixtral 8x7B as the foundation LLM and the baselines used for comparison are the same ones used in Section~\ref{sec:exp:q1}.

\begin{table*}[t]
    \tabcolsep=0.12cm
    \begin{tabular}{l|llll|llll|llll}
    \toprule
         Testing  &         \multicolumn{4}{c|}{Pet ~|~ Acc $\uparrow$} & \multicolumn{4}{c|}{Airbnb ~|~ MSE $\downarrow$}      & \multicolumn{4}{c}{Avito ~|~ RMSE $\downarrow$} \\
         ~        &             SDEdit &  DDRM & Idinvert & \method   &   SDEdit &  DDRM & Idinvert & \method    &   SDEdit &  DDRM & Idinvert & \method   \\
    \midrule
          tabular &                    0.282    &  0.269  &  0.252 & \textbf{0.338}           &         0.428 & 0.621 &  0.732  &  \textbf{0.270}     &     0.108 & 0.123 &  0.133  & \textbf{0.041}     \\
            image &                    0.285    &  0.286  &  0.267 & \textbf{0.356}           &         0.566 & 0.649 &  0.711  &  \textbf{0.486}     &     0.114 & 0.123 &  0.136  & \textbf{0.044}     \\
             text &                    0.284    &  0.284  &  0.274 & \textbf{0.349}           &         0.502 & 0.601 &  0.695  &  \textbf{0.253}     &     0.113 & 0.123 &  0.131  & \textbf{0.044}     \\
    image+tabular &                    0.307    &  0.276  &  0.256 & \textbf{0.382}           &         0.394 & 0.556 &  0.683  &  \textbf{0.251}     &     0.118 & 0.124 &  0.129  & \textbf{0.042}     \\
     text+tabular &                    0.315    &  0.306  &  0.283 & \textbf{0.377}           &         0.353 & 0.470 &  0.544  &  \textbf{0.185}     &     0.124 & 0.124 &  0.134  & \textbf{0.041}     \\
       text+image &                    0.292    &  0.286  &  0.244 & \textbf{0.378}           &         0.489 & 0.537 &  0.673  &  \textbf{0.212}     &     0.110 & 0.115 &  0.125  & \textbf{0.043}     \\
              all &                    0.334    &  0.304  &  0.281 & \textbf{0.395}           &         0.345 & 0.463 &  0.542  &  \textbf{0.178}     &     0.109 & 0.114 &  0.123  & \textbf{0.042}     \\
    \midrule
          Average &                    0.300    &  0.287  &  0.265 & \textbf{0.368}           &         0.440 & 0.557 &  0.654  &  \textbf{0.262}     &     0.112 & 0.121 &  0.130  & \textbf{0.042}     \\
    \bottomrule
    \end{tabular}
    \caption{This table presents a detailed comparison, highlighting \method's performance against embedding-based translation baselines when the model is trained on all modalities and tested on different subset modalities.}
    \label{tab:full-modality}
\end{table*}

\subsubsection{Results}
The key findings outlined in Table~\ref{tab:full-modality} underscore the superior performance of \method, which still achieves substantial gains over competing baselines across various modality combinations. These results suggest that despite no modality mismatching, our strategy holds a decisive edge over embedding-based methods. 
Specifically, on the PetFinder dataset, our technique enhances accuracy by an average of approximately 22.6\%, significantly outperforming the best-performing embedding-based methods. Similarly, in the Airbnb dataset, \method achieves a decrease of approximately 40.5\% in mean squared error, indicating a significant improvement in prediction accuracy. Moreover, in the Avito dataset, the decrease is even more pronounced, with a reduction of approximately 62.5\% in mean squared error when applying \method.

\subsection{Q3: Is Text Representation Generally More Robust Than Embedding Representation For Cross Modality Translation?}
\label{sec:exp:q3}

In this section, we aimed to understand the trade-off between performance and flexibility when converting various modalities from embedding into text, especially under modality mismatch conditions.

\subsubsection{MLLMs Baseline}
Previous experiments in Section~\ref{sec:exp:q1} and Section~\ref{sec:exp:q2} cannot compare text representation and embedding representation since converting modalities into text involves different foundation models, each with different capabilities. For a fair comparison of performance between text representation and embedding representation, the most appropriate approach is to utilize multimodal Language Model Models (MLLMs). This ensures fairness in the comparison because all modalities are converted from the same foundation model.
Therefore, we applied the following SOTA MLLMs in our experiments: Kosmos-2~\cite{peng2023kosmos} and Flamingo~\cite{alayrac2022flamingo}.
In our experiments, we leverage MLLMs as both pre-trained feature extractors and text decoders. We then employ mean pooling to aggregate representations, followed by using an MLP as a backbone model to generate predictions. We aim to compare the performance gaps between a downstream model trained on images and another trained on image captions (attributed to the dataset).

\subsubsection{Results}
Results in Table~\ref{tab:ablation:mllm} consistently reveal that downstream models trained on image captions exhibit less performance degradation compared to those trained on image embeddings in scenarios of modality mismatch. This observation holds true across all state-of-the-art multimodal LLMs we investigated. Such results strongly suggest that cross-modality translation within text representations, as facilitated by \method, proves to be a more effective and robust strategy than utilizing embedding representations when faced with modality mismatch conditions.

\begin{table}[t]
    \tabcolsep=0.25cm
    \begin{tabular}{l|rr|rr}
    \toprule
         Pet | Acc $\uparrow$ & \multicolumn{2}{c|}{Flamingo} & \multicolumn{2}{c}{Kosmos2}\\
        Test/Train & caption & image & caption & image \\
    \midrule
        text & \textbf{-0.07} & -0.10 & \textbf{-0.09} & -0.11 \\
        tabular & \textbf{-0.08} & -0.10 & \textbf{-0.12} & -0.21 \\
        text+tabular & \textbf{-0.08} & -0.11 & \textbf{-0.10} & -0.15 \\
    \bottomrule
    \toprule
        Air | MSE $\downarrow$ & \multicolumn{2}{c|}{Flamingo} & \multicolumn{2}{c}{Kosmos2}\\
        Test/Train & caption & image & caption & image \\
    \midrule
        text & \textbf{-0.00} & -0.06 & \textbf{-0.01} & -0.03\\
        tabular & \textbf{-0.03} & -0.07 & \textbf{-0.05} & -0.05\\
        text+tabular & \textbf{-0.02} & -0.07 & \textbf{-0.04} & -0.03 \\
    \bottomrule
    \end{tabular}
    \caption{Text representation shows consistently less performance degradation for cross-modality translation when explicitly compared to embedding representation. Both representations are derived from the same Multimodal LLMs for fair comparison. Nevertheless, transforming from image to caption has a slight performance reduction.}
    \label{tab:ablation:mllm}
\end{table}


\subsection{Ablation Studies}
\label{sec:exp:ablation}
This section explores the contribution of individual components within \method by conducting ablation studies. We incrementally add modules to evaluate their impact on performance, with findings summarized in Table~\ref{table:ablation}. In this section, our \method framework employs the GPT-3.5 as the foundation LLM.

\begin{table*}[t]
\tabcolsep=0.4cm
\begin{tabular}{ll|cccccc}
\toprule
    Training &       Testing &         \multicolumn{4}{c}{PetFinder ~|~ Accuracy $\uparrow$} \\
\midrule
    ~ &      ~  &  \multirow{2}{*}{SDEdit} & Text &    +Modality  &  +Reasoning  &   +Text-style \\
    ~ &      ~  & ~  & Transformation &    Summarization  &   Augmentation &   Translation \\

\midrule
    text+image     & tabular       & 0.282  & 0.310          & 0.321             & 0.338                    & \textbf{0.348} \\
    text+tabular   & image         & 0.289  & 0.329          & 0.365             & 0.363                    & \textbf{0.380} \\
    image+tabular  & text          & 0.281  & 0.305          & 0.295             & 0.321                    & \textbf{0.355} \\
    text           & image+tabular & 0.291  & 0.282          & 0.296             & 0.343                    & \textbf{0.344} \\
    text           & image         & 0.289  & 0.293          & 0.298             & 0.341                    & \textbf{0.374} \\
    text           & tabular       & 0.293  & 0.297          & 0.318             & 0.315                    & \textbf{0.357} \\
    image          & text+tabular  & 0.290  & 0.314          & 0.289             & 0.325                    & \textbf{0.341} \\
    image          & text          & 0.288  & 0.306          & 0.330             & \textbf{0.336}                    & 0.319 \\
    image          & tabular       & 0.291  & 0.300          & 0.307             & 0.303                    & \textbf{0.348} \\
    tabular        & text+image    & 0.290  & 0.194          & \textbf{0.366}             & 0.341                    & 0.360 \\
    tabular        & text          & 0.289  & 0.193          & 0.306             & 0.327                    & \textbf{0.364} \\
    tabular        & image         & 0.289  & 0.196          & 0.357             & 0.353                    & \textbf{0.364} \\
\midrule
\multicolumn{2}{c|}{Average $\pm$ Variance ($\times 10^{-4}$)} & $0.289 \pm 0.12$ & $0.277 \pm 25.91$ & $0.321 \pm 7.2$ & $0.334 \pm 2.5$ & $\textbf{0.355} \pm \textbf{2.4}$\\
\bottomrule
\end{tabular}
\caption{Ablation studies on various components of \method.
Our observations reveal that text transformations significantly enhance performance across all modality combinations except for tabular data, which is in fixed formatted text. The formatting issue is effectively solved by incorporating a summarization module, resulting in a substantial enhancement in performance. Furthermore, the inclusion of both the translation module and the reasoning augmentation module leads to further improvements in overall performance.}
\label{table:ablation}
\end{table*}

\subsubsection{Text Transformation}
Compared to the embedding-based methods SDEdit, analysis from Table~\ref{table:ablation} shows converting modality features into text enhances performance by approximately 2\%, indicating less modality mismatch during training and inference compared to embedding representations. This improvement is consistent across most data modalities, except for tabular data, which sees a decline of about 10\%. This discrepancy is attributed to the fixed format of tabular text transformation, highlighting a significant style gap with more fluid, human-like writing, particularly impacting tabular data's inference performance.

\subsubsection{Modality Summarization}
Table~\ref{table:ablation} results indicate modality summarization improves tabular data accuracy significantly from 0.277 to 0.321 on average. After this stage, \method has already outperformed the strongest competitor SDEdit. This suggests that summarization effectively standardizes text formats into a cohesive style, mitigating heterogeneity in text transformation and enhancing data format alignment.

\subsubsection{Reasoning Augmentation}

Table~\ref{table:ablation} indicates that augmentation enhanced our average performance from 0.321 to 0.334. Additionally, we have observed that it contributes to a more stable performance across different scenarios. The variance value with augmentation is substantially lower than that without it.


\subsubsection{Text-Style Translation across Modality}
According to Table~\ref{table:ablation}, text-style translation bridges training and inference phase gaps, with about 6\% improvement from 0.334 to 0.355. This enhancement is particularly notable when the gap in textual style remains consistent across phases, as seen in the image-to-table scenarios. Such consistency aids in more accurate mapping function determination by the model.

\section{Analysis and Discussion}
\label{sec:analysis}
In this section, we delve into a series of analyses and discussions, extracting valuable insights from our discoveries. 
Specifically, we provide more supportive evidence with visualization and distribution distance measurements.

\subsection{Visualization for Distribution Alignment}

\begin{figure}[t]
    \centering
    \includegraphics[width=1.0\linewidth]{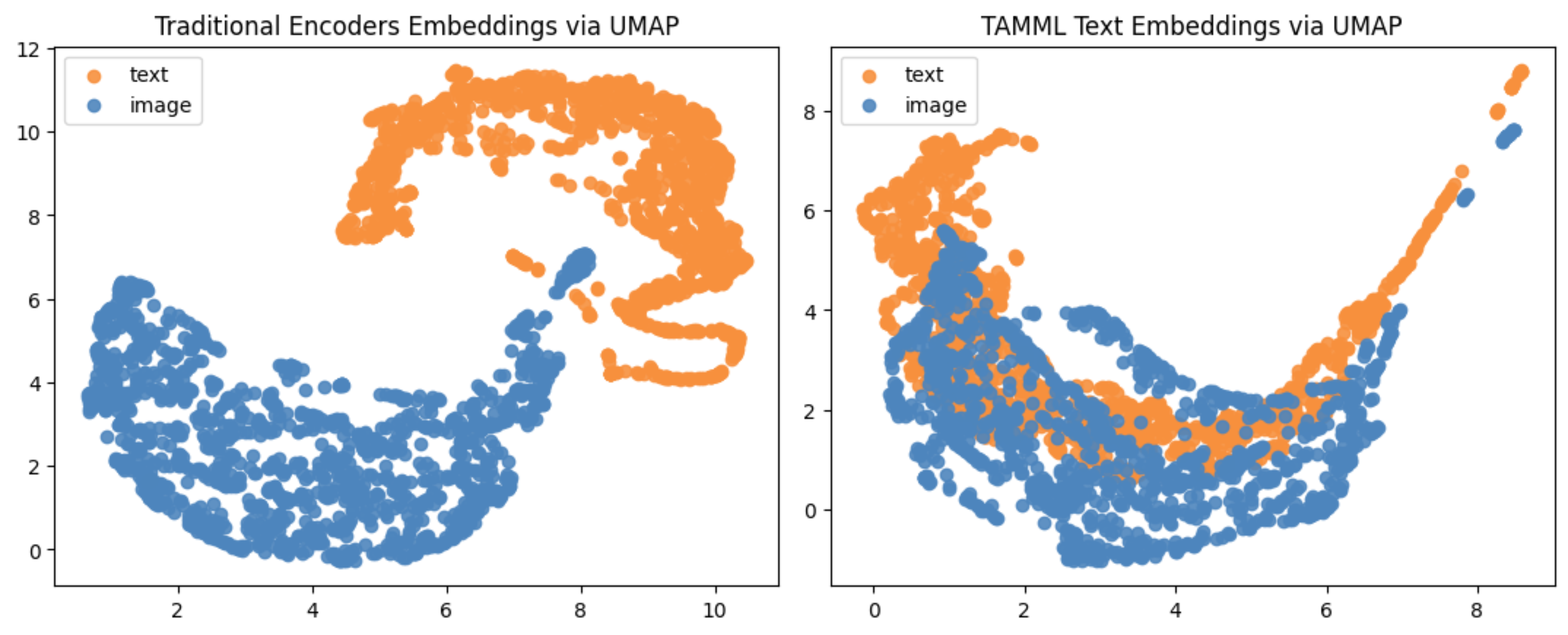}
    \caption{The left and right pictures illustrate the visualizations of embeddings for image and text data, respectively, before and after our processes.}
    \label{fig:tradition2our}
\end{figure}

In Section~\ref{sec:exp:q3}, we have validated the effectiveness of text transformation in \method through experimental performance. Furthermore, we visualized 1,400 data points in these modalities with their position-aware embeddings using UMAP~\cite{mcinnes2018umap} in  Figure.~\ref{fig:tradition2our}. The left figure illustrates the original distributions of image and text embeddings, while the right figure displays the corresponding distributions after the summarization module in \method. We observe that the distribution boundaries between image and text modalities become less distinct, which indicates they are closer in the semantic space. To be more precise, \method significantly reduces the average instance Euclidean distance between image and text in the semantic space from $10.213$ to $0.411$.

\subsection{Effects of the Image Caption Models}
Some might argue that the improvement in text transformation in \method could be attributed to the superior GPT-4 model. To investigate this, we replaced the different image caption models in our architecture with smaller open-source models. We conducted ablation studies focusing on the performance of four image foundation models. Specifically, we showed that our approach maintains strong performance even with smaller models.
Table~\ref{table:average_model_capability} showcases the results averaged across twelve training-inference modality combinations. The results suggest that using smaller image caption models does not necessarily result in significantly inferior performance with \method.

\begin{table}[t]
    \tabcolsep=0.15cm
    \begin{tabular}{l|rrrr}
    \toprule
     \multicolumn{5}{c}{\textbf{Image Caption Models}} \\
    \midrule
        Pet ~|~ Acc $\uparrow$     &  Blip2 &  Kosmos2 & Flamingo &  GPT4 \\
        \midrule
            Average  & 0.303 & 0.299 & 0.293 & \textbf{0.307}\\
        \bottomrule
    \end{tabular}
    \caption{Image caption model comparison: Each number presented here is an average derived from twelve modality combination experiments. In general, we can infer that the foundation model has only a limited impact on \method.}
    \label{table:average_model_capability}
\end{table}

\section{Conclusion and Future Directions}
\label{sec:conclusion}

Our study has effectively harnessed Large Language Models (LLMs) for multimodal learning, creating a unified semantic space that integrates various data modalities through text. Through techniques such as text transformation, text-style translation, summarization, and reasoning augmentation, we have demonstrated that operations performed in the text domain using in-context learning with LLMs can achieve comparable performance to traditional methods operating in embedding space. This approach not only opens new avenues in multimodal learning but also underscores the significant potential and advantages of text as a unifying medium.
Future efforts will focus on refining \method for broader multimodal tasks and overcoming prompt sensitivity to fully leverage text in multimodal learning.




\bibliographystyle{ACM-Reference-Format}
\bibliography{sample-base}


\clearpage
\appendix
\onecolumn

\section{Experiment Detail Setup}
\label{app:exp}

\subsection{Model Checkpoints}

We conduct all experiments with GPT-3.5-turbo as the LLM and GPT-4-vision as the image caption model through OpenAI APIs~\cite{openai}, except for the analysis experiment that compares different LLMs and foundation models.

\begin{table}[h]
\begin{tabular}{ll}
\toprule
    Model   &     Checkpoints \\
\midrule
    GPT-3.5-turbo & gpt-3.5-turbo-0613 \\
    GPT-4-vision & gpt-4-vision-preview \\
    BLIP2 & huggingface: Salesforce/blip-image-captioning-large \\
    Kosmos2 & huggingface: microsoft/kosmos-2-patch14-224\\
    Vision Transformer & huggingface: google/vit-base-patch16-224\\
    Flamingo & huggingface: openflamingo/OpenFlamingo-9B-vitl-mpt7b\\
    Longformer & huggingface: allenai/longformer-base-4096\\
    LLAMA-2-7b-chat & huggingface: meta-llama/Llama-2-7b-chat \\
    LLAMA-2-13b-chat & huggingface: meta-llama/Llama-2-13b-chat\\
    LLAMA-2-70b-chat & huggingface: meta-llama/Llama-2-70b-chat\\
    Mixtral-8x7b & huggingface:mistralai/Mixtral-8x7B-Instruct-v0.1\\

\bottomrule
\end{tabular}
\caption{Model checkpoints.}
\label{tab:checkpoint}
\end{table}

\subsection{Hyperparameters}

\begin{table}[h]
\begin{tabular}{ll}
\toprule
    Model   &     Hyperparameters \\
\midrule
    GPT-3.5-turbo & temperature=1, max\_tokens=4096  \\
    GPT-4-vision & temperature=0.8, max\_tokens=300\\
    BLIP2 & default parameter\\
    Kosmos2 & default parameter\\
    Vision Transformer & default parameter\\
    Flamingo & default parameter\\
    Longformer & max\_length=2048\\
    LLAMA-2-7b-chat & temperature=1, max\_tokens=4096 \\
    LLAMA-2-13b-chat & temperature=1, max\_tokens=4096 \\
    LLAMA-2-70b-chat & temperature=1, max\_tokens=4096 \\
    Mixtral & temperature=1, max\_tokens=4096 \\
    SDEdit & 
    batch\_size=1, sample\_step=3, noise\_scale=150\\
    DDRM & batch\_size=1, degredation\_type=deno, noise=1.5\\
    Idinvert & batch\_size=64, gradient\_accumulate=8, network\_capacity=32\\

\bottomrule
\end{tabular}
\caption{Hyper parameters.}
\label{tab:hyperparameter}
\end{table}

\subsection{Dataset}
\label{app:exp:dataset}

\begin{table}[h]
    \begin{tabular}{ll}
        \toprule
            \multicolumn{2}{c}{PetFinder} \\
        \midrule
            Field   &     Value \\
        \midrule
            url & \url{https://www.kaggle.com/competitions/petfinder-adoption-prediction} \\
            $\#$ instances & 13453 \\
            tabular columns & 23 \\
        \bottomrule
    \end{tabular}
\end{table}
\begin{table}[h]
    \begin{tabular}{ll}
        \toprule
            \multicolumn{2}{c}{Airbnb} \\
        \midrule
            Field   &     Value \\
        \midrule
            url & \url{ http://insideairbnb.com/get-the-data/} \\
            $\#$ instances  & 12184\\
            tabular columns & 30\\
        \bottomrule
    \end{tabular}
\end{table}
\begin{table}[h]
    \begin{tabular}{ll}
        \toprule
            \multicolumn{2}{c}{Avito} \\
        \midrule
            Field   &     Value \\
        \midrule
            url & \url{ https://www.kaggle.com/competitions/avito-demand-prediction/data} \\
            $\#$ instances  & 7000\\
            tabular columns & 18\\
        \bottomrule
    \end{tabular}
\caption{Dataset Meta Info}
\label{tab:dataset}
\end{table}

\subsection{Foundation Models}
\label{app:exp:foundation}
For image modality, we utilize the embedding layer and tokenization method of the Vision Transformer~\cite{dosovitskiy2010image}. This process splits the image into fixed-size patches and then projects each patch to obtain embeddings. For tabular modality, we employ the FT-Transformer\cite{gorishniy2021revisiting} method to encode, dividing tabular features into numeric and categorical with separate projection layers for dimension enhancement. For text modality, the embedding layer of Longformer\cite{beltagy2020longformer} is used for projection.



\clearpage

\section{Analysis and Discussion}
\label{app:analysis}

\subsection{In-context Modality Transfer Outperforms Zero-shot Learning Based Methods}
\label{app:analysis:2}

Text-style translation across modalities in \method transforms the training modality combination into the testing modality combination to reduce the semantic gap between them using LLMs.
Similar concepts are used in zero-shot learning baselines, which create a generative model for modality translation.
For comparison, we collected different pairs of training and testing data and created visualizations for each one of them.

\begin{figure}[h]
    \centering
    \includegraphics[width=1.0\linewidth]{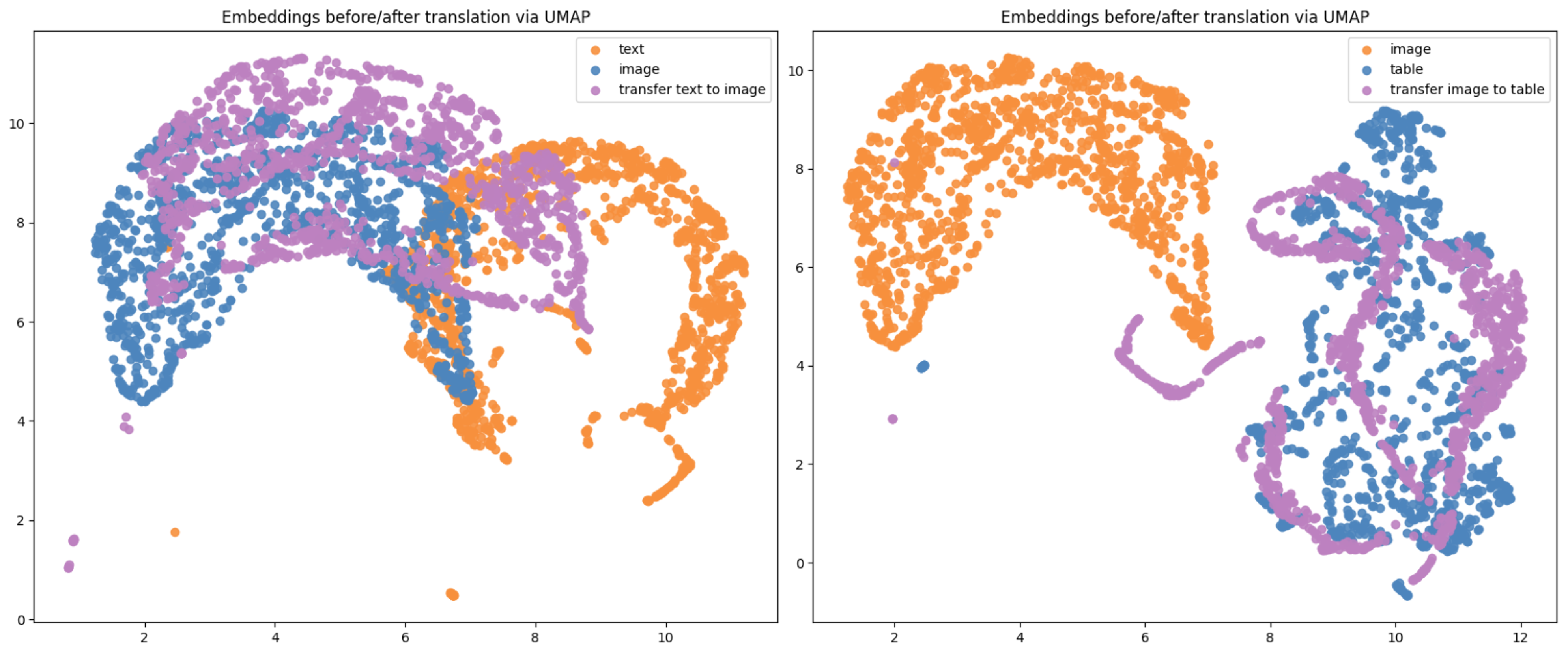}
    \caption{Cross Modality Translation (Ours): training data map to the distribution of target modality.}
    \label{fig:translation-emb}
\end{figure}

\begin{figure}[h]
    \centering
    \includegraphics[width=1.0\linewidth]
    {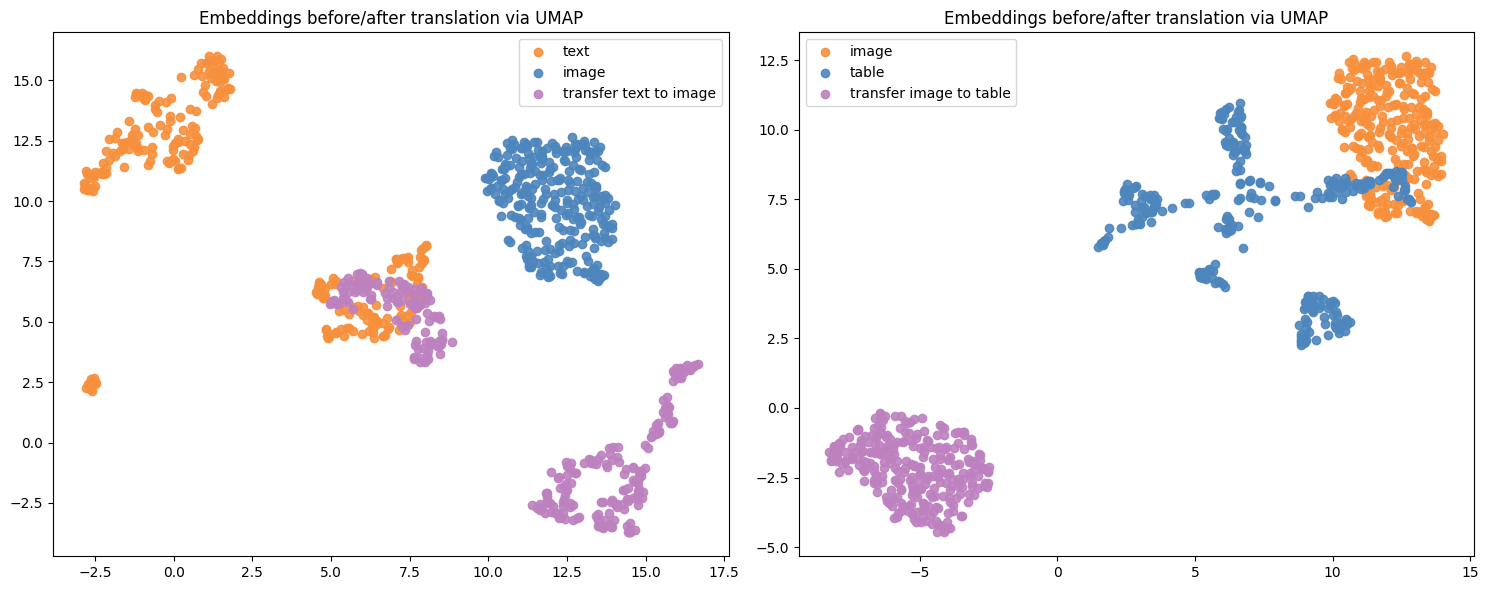}
    \caption{Cross Modality Translation (PetFinder / SDEdit): training data map to the distribution of target modality.}
    \label{fig:translation-emb-sd-pet}
\end{figure}

Orange is the source modality, blue is the target modality, and purple is the source modality after transformation.
Visualization results of Ours are shown in Figure~\ref{fig:translation-emb}.
Visualization results of SDEdit are shown in Figure~\ref{fig:translation-emb-sd-pet}.
As the results indicate, our translation effectively maps to closely align with the target modality in semantic space.


\clearpage
\section{Extra Experiment Results}
\label{app:result}

In this section, we try to answer additional inquiries, Q4 and Q5, where Q4 explores the performance of text-based solutions versus embedding-based solutions when training and testing modalities are identical, and Q5 compares our zero-shot in-context learning approach to non-zero-shot methods, such as domain adaptation.

\subsection{Q4: Text-based Solutions Versus Embedding-based Solutions When Training And Testing Modalities Are Identical}

Table~\ref{tab:main-extra} provides the experiment results under no train/test modality mismatch.

\begin{table}[h]
\tabcolsep=0.2cm
\begin{tabular}{lll}
\toprule
     Train \& Test &    Regular (Embedding) &  \method (Text) \\
\midrule
    text          &           0.352 & \textbf{0.382} \\
    image         &           0.273 & \textbf{0.369} \\
    tabular       &  \textbf{0.429} &          0.394 \\
    text+image    &           0.286 & \textbf{0.400} \\
    text+tabular  &  \textbf{0.411} &          0.404 \\
    image+tabular &           0.403 & \textbf{0.408} \\
         
\bottomrule
\end{tabular}
\caption{Experiment results under no train/test modality mismatch condition. Under this condition, \method does not show performance degradation and even performs better in several modality combinations. The regular method means the downstream model is trained on embedding representations. Note that this result differs from the result in Table~\ref{tab:ablation:mllm} because the foundation models used for generating embedding and text representations are not the same.}
\label{tab:main-extra}
\end{table}

\subsection{Q5: How does \method compare to non-zero-shot methods?}

Table~\ref{tab:main-extra-setting} provides the experiment results under modality mismatch with different test time finetuning settings (not zero-shot). The settings are as follows:
\begin{itemize}
    \item no finetuning: complete mismatch scenario same as main result experiments.
    \item unsupervised domain adaptation: finetune- the downstream model given the information of inference modality but without labels. We adopted the ADDA~\cite{tzeng2017adversarial} method.
    \item supervised training (with all modalities): the downstream model given the information of paired train/inference time modality with labels. This means that the modality used in testing is fully trained.
\end{itemize}

\begin{table}[h]
\tabcolsep=0.2cm
\begin{tabular}{ll|llll}
\toprule
     Train & Test &    no finetuning: Emb & no finetuning: \method &  unsupervised domain adaptation    & supervised training (all modalities) \\
\midrule
    text         &     image    &   0.288   &   \textbf{0.374}  &     0.195   & 0.338 \\
    text         &   tabular    &   0.289   &   0.357  &     0.281   & \textbf{0.359} \\
    image        &      text    &   0.270   &   \textbf{0.319}  &     0.276   & 0.306 \\
    image        &   tabular    &   0.273   &   0.348  &     0.276   & \textbf{0.359} \\
    tabular      &      text    &   0.289   &   \textbf{0.364}  &     0.195   & 0.306 \\
    tabular      &     image    &   0.279   &   \textbf{0.364}  &     0.195   & 0.338 \\

\bottomrule
\end{tabular}
\caption{The experiment results showed a condition with other non-zero-shot methods. Under this condition, \method shows no performance degradation and even performs better in several modality combinations in zero-shot.}
\label{tab:main-extra-setting}
\end{table}

\end{document}